\documentclass{article}
\usepackage{xcolor}
\usepackage{cite}
\usepackage{amsmath}
\usepackage{amssymb}
\usepackage[preprint]{corl_2026}
\usepackage{ifthen}

\definecolor{SaddleBrown}{RGB}{139,69,19}
\definecolor{Purple}{RGB}{102,0,153}

\usepackage{graphicx}
\usepackage{subfig}
\usepackage[utf8]{inputenc}
\usepackage{pgfplots}
\usepackage{float}
\usepackage{booktabs}
\usepackage{multirow}
\usepackage{bm}
\usepackage{algorithm}
\usepackage{algpseudocode}
\usepackage{soul}
\usepackage{textcomp}
\usepgfplotslibrary{groupplots,dateplot}
\usetikzlibrary{patterns,shapes.arrows,arrows.meta}
\pgfplotsset{compat=newest,scaled x ticks=false}

\usepackage{mwe}
\usepackage{tikz}
\usetikzlibrary{external,positioning,arrows.meta,bending,calc}
\usepgfplotslibrary{external}
\usepackage[capitalise]{cleveref}
\usepackage{wrapfig}
\usepackage[per-mode=symbol]{siunitx}

\DeclareMathOperator*{\argmin}{arg\,min}

\newcommand{\vp}{\mathbf{p}}                       
\newcommand{\vc}{\mathbf{c}}                       
\newcommand{\vh}{\mathbf{h}}                       
\newcommand{\va}{\mathbf{a}}                       
\newcommand{\vb}{\mathbf{b}}                       
\newcommand{\zhat}{\hat{\mathbf{z}}}               
\newcommand{\dhat}{\hat{\mathbf{d}}}               
\newcommand{\norm}[1]{\left\lVert #1 \right\rVert}

\newcommand{\vatcmd}{\mathbf{a}_t^{\mathrm{cmd}}}
\newcommand{\vad}{\mathbf{a}_d}
\newcommand{\vanet}{\mathbf{a}_{\mathrm{net}}}

\newcommand{\Rtilt}{\mathbf{R}_{\mathrm{tilt}}}

\newcommand{\sidelength}{s_{\text{box}}} 

\newcommand{\nenv}{N_{\text{env}}}              
\newcommand{\nupd}{N_{\text{upd}}}              
\newcommand{\maxsteps}{T_{\text{ep}}}           
\newcommand{\nseed}{N_{\text{seed}}}            
\newcommand{\captureradius}{c_r}                
\newcommand{\escaperadius}{e_r}                

\newcommand{\lamvalign}{\lambda_{\text{align}}}      
\newcommand{\lamvclose}{\lambda_{\text{closing}}}    
\newcommand{\lamacc}{\lambda_{\text{acc}}}             
\newcommand{\lamjerk}{\lambda_{\text{jerk}}}           
\newcommand{\lamvmax}{\lambda_{\text{vmax}}}           
\newcommand{\lamyaw}{\lambda_{\text{yaw}}}             

\newcommand{\horizon}{H}                        
\newcommand{\lrA}{\eta_{a}}                     
\newcommand{\lrC}{\eta_{c}}                     
\newcommand{\lrfinal}{\eta_{\text{final}}}      
\newcommand{\lrdecay}{\rho_{\text{lr}}}         
\newcommand{\wdecay}{w_{d}}                     
\newcommand{\gmax}{g_{\max}}                    
\newcommand{\discount}{\gamma}                  
\newcommand{\Kepoch}{K}                         
\newcommand{\Nmb}{N_{\text{mb}}}                
\newcommand{\gaelam}{\lambda_{\text{GAE}}}      
\newcommand{\ppoclip}{\varepsilon}              
\newcommand{\cval}{c_{v}}                       
\newcommand{\cent}{c_{\text{ent}}}              

\newcommand{\vtgt}{v_{\mathrm{tgt}}}            
\newcommand{\tiltrpy}{\boldsymbol{\theta}_{\text{tilt}}}  
\newcommand{\axisrad}{r_{\text{axis}}}          
\newcommand{\aratio}{\rho_{\text{ar}}}          
\newcommand{\zrate}{z_{\mathrm{rate}}}          

\newcommand{\dtsim}{\Delta t_{\text{sim}}}      
\newcommand{\dtctrl}{\Delta t_{\text{ctrl}}}    
\newcommand{\mass}{m}                           
\newcommand{\kdone}{k_{d,1}}                    
\newcommand{\kdtwo}{k_{d,2}}                    

\newcommand{\saccel}{\beta_{a}}                 
\newcommand{\waccel}{\tau_{a}}                  
\newcommand{\daccel}{\delta_{a}}                

\newcommand{\syaw}{\beta_{\psi}}                
\newcommand{\wyaw}{\tau_{\psi}}                 
\newcommand{\dyaw}{\delta_{\psi}}               

\newcommand{\svel}{\beta_{v}}                   
\newcommand{\wvel}{\tau_{v}}                    

\newcommand{\grav}{g}                           
\newcommand{\Ixx}{I_{xx}}                       
\newcommand{\Iyy}{I_{yy}}                       
\newcommand{\Izz}{I_{zz}}                       
\newcommand{\Ixy}{I_{xy}}                       
\newcommand{\Iyz}{I_{yz}}                       
\newcommand{\Ixz}{I_{xz}}                       
\newcommand{\armlen}{\ell_{\text{arm}}}         
\newcommand{\momentscale}{k_{\text{moment}}}          
\newcommand{\motorangle}{\theta_{m}}            
\newcommand{\rotortc}{\tau_{r}}                 
\newcommand{\rpmmin}{\Omega_{\min}}             
\newcommand{\rpmmax}{\Omega_{\max}}             

\title{Learning Agile Intruder Interception using Differentiable Quadrotor Dynamics}

\author{
  Michael Anoruo$^{\dagger,*}$, Xiaoyu Tian$^{\dagger,*}$, Abhishek Rathod$^{\dagger}$, Timothy Naudet$^{\S}$\\
  \textbf{Thomas Canchola$^{\S}$, Eric Sturzinger$^{\S}$, Kshitij Goel$^{\dagger}$, and Wennie Tabib$^{\dagger}$}\\
  $^{\dagger}$Carnegie Mellon University, $^{\S}$Artificial Intelligence Integration Center\\
  \url{http://rislab.github.io/projects/catchrl.html}
}

\begin{document}
\maketitle
{\let\thefootnote\relax\footnotetext{$^*$Equal contribution.}}


\begin{abstract}
This paper presents a methodology for learning a control policy to intercept
an intruder using the 3D direction unit vector to the intruder and the
interceptor state. Prior deep reinforcement learning approaches assume either
relative position or distance to the intruder is available, but this information
is not readily accessible in real-world applications that employ passive,
monocular camera sensors. Instead, we propose a
solution that leverages an analytical policy gradient method using differentiable quadrotor dynamics
to learn agile interception at speeds up to
\SI{10}{\meter\per\second}. The proposed approach outperforms baseline methods
that utilize simplified point mass dynamics by an average of 30\%.
\end{abstract}



\begin{figure}[H]
  \begin{center}
    \includegraphics[trim=0 85 0 95,clip,width=\textwidth]{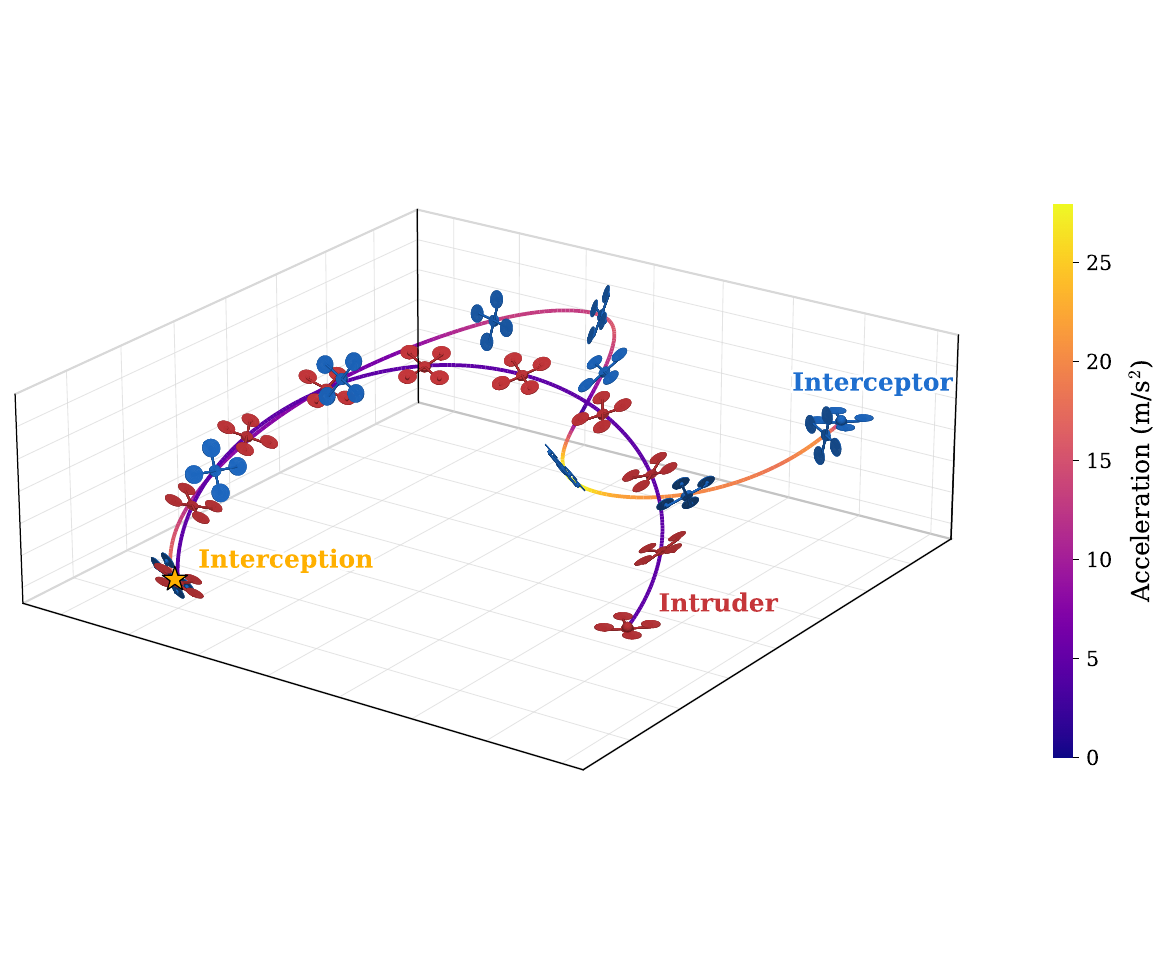}
  \end{center}
  \caption{\label{fig:glory-shot}Plot of a rollout in simulation using the
  proposed control policy. The interceptor (blue) aggressively pursues the
  intruder (red) flying at \SI{5.8}{\meter\per\second} and collides with it at
  the star (yellow) labeled interception.  The acceleration along the
  trajectories taken by each agent is plotted from blue (low acceleration) to
  yellow (high acceleration). Both vehicles are assumed to operate in open air
  environments in the absence of obstacles.}
\end{figure}

\section{Introduction}
Agile unmanned aerial systems have the potential to disrupt airport
operations and damage critical urban infrastructure
\cite{skraparlis2024novel}. The ubiquity and
low-cost of commercial off-the-shelf unmanned aerial systems pose
challenges for defending restricted airspaces. Thus, the demand for
cost-effective countermeasures against unauthorized drone activity
is increasing~\cite{gavin2026agile}. In this work,
we propose a solution that intercepts an intruder drone
through forced collisions. Prior works leveraging Deep Reinforcement Learning
(DRL) for interception require relative 3D position or distance
information during inference for interception; however, this
information is not readily available in real-world
applications. Instead, we assume the interceptor is equipped with
multiple cameras whose images provide 360$^{\circ}$
coverage. Given this sensing model,
object detection is possible on a unit sphere around the interceptor, which
determines the direction of the intruder relative to the interceptor.

The problem we address is how to train a policy that uses only the
unit direction vector to the intruder and interceptor state as inputs
during inference.
To this end, we make the following contributions: (1) an end-to-end
control policy trained via differentiable simulation for interception;
(2) extensive evaluations in simulation that compare the approach
against state-of-the-art RL algorithms and multiple dynamics models;
and (3) an open source software release\footnote{\url{https://github.com/rislab/catchrl}}.

\section{Related Work}
This section provides an overview of prior work most closely aligned
with our approach. These works can be broadly categorized as
autonomous interception and learning-based navigation.

\textbf{Interception:} Prior works have leveraged DRL for drone
interception~\cite{gavin2026agile,roncero2025learned,zheng2025non,pierre2023multi,logiewa2023dynamic};
however, these approaches assume that the intruder's 3D position relative
to the interceptor is known during inference, which is challenging
to estimate using passive monocular sensing. Our approach assumes
that only the 3D direction vector without distance to the intruder is
known. In practice, this information could be obtained from an object
detector (e.g. YOLO~\citep{redmon2016you}) through a set of onboard cameras.

LiDAR-based intruder detection and interception has been demonstrated
in~\cite{pliska2024safe,vrba2025onboard}, but LiDARs are active sensors that
emit pulses of infrared laser light~\cite{ryde2009performance}, which makes them
detectable~\cite{zygmunt2020laser} compared to passive sensors like RGB cameras.
LiDARs are also relatively expensive, so methods that leverage passive
sensors are preferred.~\citet{yan2025precise,guo2026dynamic} develop visual
servoing for intercepting a dynamic intruder, but the approaches are
evaluated using simple trajectories that vary the $z$-altitude exclusively or move
along the $xy$-plane. In contrast, we evaluate our approach with agile 3D
trajectories. \citet{liu2025autonomous} have proposed using noisy distance
observations via a microphone for interception but require two
communicating vehicles to intercept one intruder. We propose a strategy that
requires one interceptor per intruder.

Prior works have also considered how jamming could be used to
interfere with GNSS receivers to intercept
drones~\cite{souli2023multi,valianti2024cooperative,ma2025lure,souli2025enhanced};
however, this is ineffective for systems that can operate in
GPS-denied environments (e.g., using visual inertial odometry instead
of GPS to navigate). Nets have also been considered to catch rogue
drones~\cite{rothe2025autonomous}, but multiple systems are
required to coordinate to transport the net, which necessitates
precise localization using a differential GNSS solution. Further, the
solution requires external target detection and tracking~\cite{rothe2025autonomous}, which limits
real-world deployment.

\textbf{Learning-based Quadrotor Navigation using Differentiable Physics:}
Learning-based quadrotor navigation methods using differentiable quadrotor
physics models~\citep{zhang2025learning,lee2026quadrotor,li2026simple} have
demonstrated performance gains and improved sample efficiency over imitation
learning methods~\citep{loquercio2021learning}.~\citet{zhang2025learning}
contribute a neural policy that processes a depth image and quadrotor state 
to output acceleration commands. Instead of leveraging a critic network, their
method relies on a simplified point mass dynamics model through which gradients
can flow during backpropagation. Extensions to this work have been
proposed towards improving yaw control~\citep{lee2026quadrotor} and training
stability~\citep{li2026simple}. To the best of our knowledge, differentiable
simulation has not been used in prior work to address the interception problem.
We address this gap in this paper by developing an end-to-end interception policy
trained via differentiable simulation.

\section{Methodology}
We design a learning-based control policy for a quadrotor agent intercepting a
dynamic intruder in an open 3D workspace. The policy (see~\cref{fig:net-diag})
receives quadrotor proprioception and a unit direction vector to the intruder.
A recurrent neural network (RNN) maps the onboard observations to a 3D thrust
and heading command which are used to intercept the intruder.

At time $t$, let the state of this system be $x_t \in \mathcal{X}$,
where $\mathcal{X}$ is the continuous state space for the interceptor and the
intruder. The interceptor observes $o_t = \bigl[\mathbf{v}_t,
\mathbf{R}_t, \dhat_t\bigr]$, where $\mathbf{v}_t \in \mathbb{R}^3$
is the interceptor's linear velocity, $\mathbf{R}_t \in \mathrm{SO}(3)$ is the
interceptor body-to-start
rotation matrix, and
$\dhat_t \in \mathbb{S}^2$ is the unit direction vector to the intruder from the
interceptor, expressed in the frame
aligned with the starting pose of the interceptor. The policy outputs $u_t
= \bigl[\vatcmd, \psi_t\bigr]$ consist of the 3D commanded
mass-normalized thrust vector for the interceptor $\vatcmd
\in \mathbb{R}^{3}$ and a scalar yaw angle $\psi_t \in \mathbb{R}$.
These outputs are tracked using a PD attitude controller~\cite{mellinger2012trajectory}
onboard the quadrotor interceptor. In general, all the inputs and outputs of
the policy are expressed in the start frame.

The policy $u_t = \pi_{\theta}(o_t)$ is represented using a neural network with $\theta$
parameters.  We leverage the Analytical Policy Gradient (APG) method~\cite{wiedemann2023training} which
backpropagates gradients through time (BPTT) using a differentiable dynamics
model to optimize the network parameters via gradient
descent~\citep{zhang2025learning}. Formally, we want to find the optimal policy
parameters that minimize the loss over a trajectory containing $T$ timesteps,
$\theta^* = \argmin_{\theta} \sum_{t = 0}^{T} \mathcal{L}(x_t,
\pi_{\theta}(o_t))$,
where $\mathcal{L}$ is the per-step loss. To this end, the
following sections describe the network
architecture for $\pi_{\theta}$, the differentiable dynamics model used for APG,
and the objective functions that contribute to $\mathcal{L}$ and enable dynamic intruder
interception.

\subsection{Policy Network Architecture\label{sec:network}}
\begin{figure}[t]
    \centering
    \includegraphics[trim=0 250 0 310,clip,width=\linewidth]{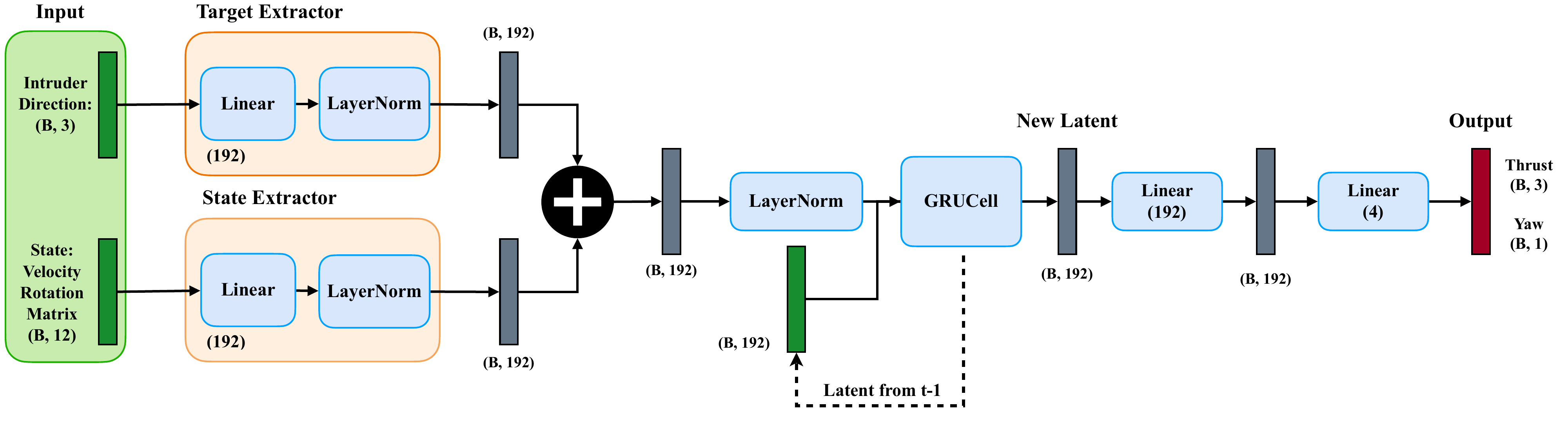}
    \caption{\label{fig:net-diag}Overview of the network architecture
        for the interception control policy. The 3D direction vector to the intruder
        from the interceptor is used as an input to the policy
        along with the interceptor linear velocity and rotation matrix flattened to
        a 9-dimensional vector. MLP encoders process the intruder and
        interceptor data separately to produce 192-dimensional embeddings, which
        are summed and passed to a GRU with 192 hidden units. The output of the GRU
        is fed through a single hidden layer before a linear head to generate the
        interceptor control commands.}
\end{figure}
The policy network is designed to exploit the temporal structure of the
interception task while remaining compatible with BPTT-based optimization~\citep{zhang2025learning}
(see~\cref{fig:net-diag}). The proprioceptive state $[\mathbf{v}_t, \mathbf{R}_t]$
and the 3D unit direction vector to the intruder $\dhat_t$ are processed by two separate MLP
encoders, each producing a 192-dimensional embedding. The two embeddings are
summed and passed to a Gated Recurrent Unit (GRU) with 192 hidden units, whose
output is fed through a single hidden layer (192 units) before a linear head
emitting the four-dimensional action $[\vatcmd, \psi_t]$.
The recurrent state enables the policy to integrate information across timesteps,
implicitly estimating intruder velocity and acceleration from the temporal
sequence of 3D direction measurements. This is essential for predictive
interception given that the intruder velocity and absolute position are unobservable
during inference.

\subsection{Differentiable Dynamics Model}
Recent works that leverage APG and BPTT for training a quadrotor navigation policy
opt for a simplified point mass dynamics model to enable higher sample efficiency
and shorter training times~\citep{zhang2025learning,lee2026quadrotor}.
However, for the dynamic interception task, we found that the nonlinear quadrotor dynamics model
provides a higher interception accuracy relative to the point mass approximation.

The quadrotor is modeled as a rigid body with state
$[\mathbf{p}, \mathbf{v}, \mathbf{q}, \boldsymbol{\omega}]$
representing position, linear velocity, attitude quaternion, and body-frame
angular velocity, respectively. Its dynamics evolve according to
\begin{equation}
    \dot{\mathbf{p}} = \mathbf{v},
    \qquad
    \dot{\mathbf{v}} = \tfrac{1}{\mass}\mathbf{R}_{wb}\mathbf{(f + d(v))} - \grav\zhat,
\end{equation}
\begin{equation}
    \dot{\mathbf{q}} = \tfrac{1}{2}\, \mathbf{q} \otimes \boldsymbol{\omega},
    \qquad
    \dot{\boldsymbol{\omega}} = \mathbf{J}^{-1}\!\bigl(\boldsymbol{\tau} - \boldsymbol{\omega} \times \mathbf{J}\boldsymbol{\omega}\bigr),
\end{equation}
where $\mathbf{R}_{wb}$ is the body-to-world rotation, $\mathbf{f} = [0, 0, u_1]^\top$
is the collective thrust in the body frame, $\boldsymbol{\tau}$ is the body
torque, $\mathbf{J}$ is the inertia matrix, and $\grav$ is gravity.
The aerodynamic drag is modeled as
\begin{equation}
    \mathbf{d}(\mathbf{v}) = -\kdone \|\mathbf{v}\| \mathbf{v} - \kdtwo \mathbf{v},
    \label{eq:drag_model}
\end{equation}
where $\kdone$ and $\kdtwo$ are drag coefficients.
The collective thrust $u_1$ and body torque $\boldsymbol{\tau}$ are produced
by four rotors. An attitude-rate controller converts the
policy's high-level thrust-and-yaw command into desired rotor speeds $\Omega_{\mathrm{cmd}, i}$.
To capture real-world latency, the actual rotor speeds $\Omega_i$ are modeled with a first-order delay:
\begin{equation}
    \dot{\Omega}_i = \frac{1}{\tau_r} (\Omega_{\mathrm{cmd}, i} - \Omega_i),
\end{equation}
where $\tau_r$ is the rotor time constant. The forces and torques are then
generated via a quadratic thrust model and mixer matrix. The
ordinary differential equation above is integrated with the fixed-step 4th order
Runge--Kutta (RK4) method. This model is differentiable and can be used within
the APG-based policy optimization.

\subsection{Objective Functions}
The training objective minimizes the weighted sum of the guidance and
regularization terms. The total per-step loss is
\begin{equation}
    \mathcal{L}_t = \lambda_{\mathrm{align}}\mathcal{L}_{\mathrm{align}}
                + \lambda_{\mathrm{close}}\mathcal{L}_{\mathrm{close}}
                + \lambda_{\mathrm{acc}}\mathcal{L}_{\mathrm{acc}}
                + \lambda_{\mathrm{jerk}}\mathcal{L}_{\mathrm{jerk}}
                + \lambda_{\mathrm{vmax}}\mathcal{L}_{\mathrm{vmax}}
                + \lambda_{\mathrm{yaw}}\mathcal{L}_{\mathrm{yaw}}.
                \label{eq:objective_fn}
\end{equation}
Each term is described below. We leverage the intruder's position and
velocity as privileged information for loss computation. This information is not
provided as an input to the policy during deployment. This setup encourages
emergent interception behavior while maintaining realistic observation
constraints at inference time. All terms are calculated in the world frame.
\begin{wrapfigure}[9]{r}{0.45\textwidth}
  \centering
  \includegraphics[trim=2.2cm 1.75cm 3cm 0.5cm, clip,width=0.43\textwidth]{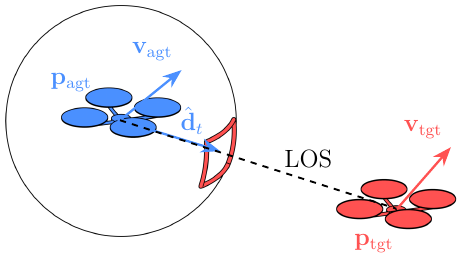}
  \label{fig:los}
\end{wrapfigure}

\paragraph{Guidance:}
Inspired by parallel navigation~\citep{yuan1948homing}, these terms
enable intruder interception (see inset figure). Let $\mathbf{d}_t =
\mathbf{p}_{\mathrm{tgt}} - \mathbf{p}_{\mathrm{agt}}$
be the relative position
and $\mathbf{v}_{\mathrm{rel},t} = \mathbf{v}_{\mathrm{tgt}} -
\mathbf{v}_{\mathrm{agt}}$ the relative velocity of the interceptor
($\mathrm{agt}$) and the intruder ($\mathrm{tgt}$). The unit line-of-sight (LOS)
vector is $\dhat_t = \mathbf{d}_t/\|\mathbf{d}_t\|$.
The parallel-navigation criterion decomposes into two scalar
quantities. The first term captures the angular drift of the LOS,
\begin{equation}
    m_t = \mathcal{L}_{\mathrm{align}} = \bigl\|\dhat_t \times \hat{\mathbf{v}}_{\mathrm{rel},t}\bigr\|
        = \sin \angle\bigl(\mathbf{d}_t, \mathbf{v}_{\mathrm{rel},t}\bigr),
\end{equation}
which vanishes when the relative velocity is aligned with the LOS. The
second term captures the rate of approach along the LOS,
\begin{equation}
    v_{c, t} = -\mathcal{L}_{\mathrm{close}} = -\dhat_t^{\!\top} \mathbf{v}_{\mathrm{rel},t},
\end{equation}
which is positive when the gap is shrinking. The policy is trained to
\emph{minimize} $m_t$ and \emph{maximize} $v_{c,t}$. The
alignment term penalizes the sine of the angle between the relative position and
relative velocity vectors, vanishing when the agent moves along the
LOS to the intruder. The closing term (note the sign --- minimizing the loss
\emph{maximizes} closing speed) drives the agent to shrink the gap
rather than pointing to the intruder.

\paragraph{Regularization:}
To produce smooth trajectories, we penalize commanded acceleration,
jerk, and speed:
\begin{equation}
    \mathcal{L}_{\mathrm{acc}} = \|\mathbf{a}_t\|^2, \qquad
    \mathcal{L}_{\mathrm{jerk}} = \|\mathbf{j}_t\|^2, \qquad
    \mathcal{L}_{\text{vmax}} = \max\bigl(0,\, \|\mathbf{v}_t\| - v_{\mathrm{des}}\bigr)^2.
\end{equation}
The acceleration penalty discourages excessive control effort; the jerk
penalty enforces temporal smoothness, mitigating high-frequency oscillations
that destabilize BPTT; and the one-sided velocity penalty keeps trajectories
within the platform's flight envelope while leaving the agent free to operate
below the desired speed $v_{\mathrm{des}}$.

\textbf{Yaw Alignment:} We add a loss term that helps the policy align the body
$x$-axis with the intruder direction:
\begin{equation}
    \mathcal{L}_{\mathrm{yaw}} = -\dhat_t^{\!\top} \hat{\mathbf{b}}_{1,t},
\end{equation}
where $\hat{\mathbf{b}}_{1,t}$ is the world-frame projection of the body
x-axis onto the horizontal $xy$-plane.

\section{Results}
\begin{figure}
    \subfloat[Ellipse]{\label{fig:traj_ellipse}\includegraphics[width=0.32\linewidth]{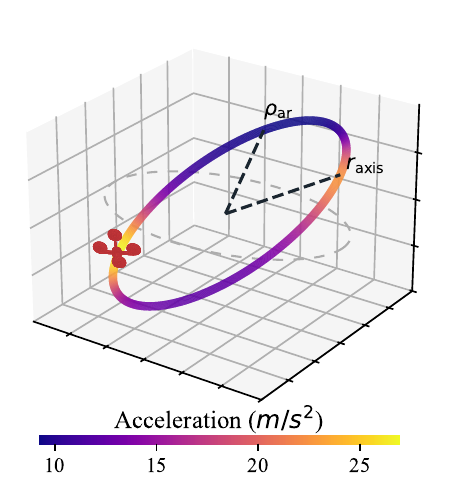}} \hfill
    \subfloat[Spiral]{\label{fig:traj_spiral}\includegraphics[width=0.32\linewidth]{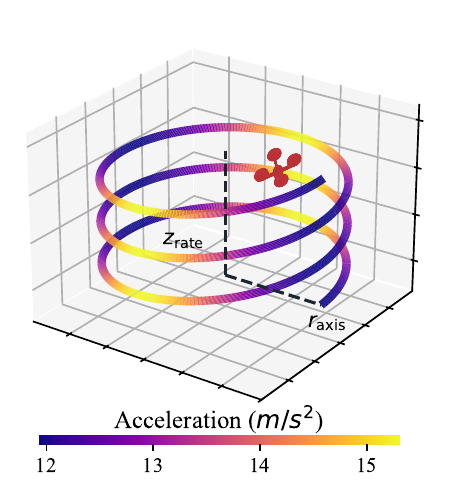}} \hfill
    \subfloat[Lemniscate]{\label{fig:traj_lemniscate}\includegraphics[width=0.32\linewidth]{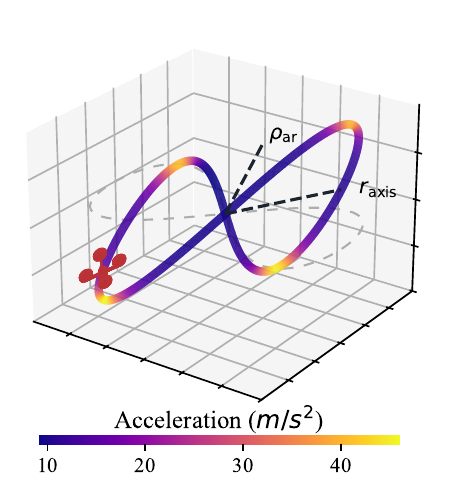}}
    \caption{\label{fig:trajectories} Example~\protect\subref{fig:traj_ellipse}
        ellipse,~\protect\subref{fig:traj_spiral} spiral,
        and~\protect\subref{fig:traj_lemniscate} lemniscate intruder
        trajectories. The ellipse trajectories are used for training
        while all are used in evaluation. The parameters of the trajectories
        are randomly sampled (\cref{appx-ssec:intru-traj}).  A subset of these parameters,
        $\axisrad$, $\aratio$, and $\zrate$, are visualized and denote the
        semi-axis, aspect ratio, and vertical ascent
        rate for the spiral, respectively.  The trajectories are colorized
        according to the magnitude of the
        acceleration.~\protect\subref{fig:traj_ellipse}--\protect\subref{fig:traj_lemniscate}
        present increasingly challenging interception tasks.
    }
\end{figure}

\textbf{Experimental Design:}
Our method is evaluated using a custom-built parallelized, GPU-accelerated
simulator written in PyTorch~\citep{paszke2019pytorch}.
The intruder is successfully intercepted if its position
is within a capture radius $\captureradius$ at any time. If after
$\maxsteps$ timesteps or when the distance between the
intruder and interceptor exceeds the escape radius $\escaperadius$,
the interception is considered to have failed.
Each timestep is $\dtctrl$ seconds long.
The success rate is measured as the number of rollouts that terminate in success
divided by the total number of rollouts. For each rollout, the
interceptor's initial position and orientation are randomized within a
box of side length $\sidelength$, and the intruder is initialized on a
sampled trajectory. All evaluation rollouts utilize the nonlinear quadratic
dynamics model. We sweep loss coefficients of our approach as detailed
in~\cref{app:paramsweep}.
The values for all parameters are provided in~\cref{app:final_params}.

The sampled trajectories consist of three types: ellipse,
spiral, and lemniscate (see~\cref{fig:trajectories}). Ellipse
trajectories are used for both training and evaluation, whereas the
lemniscate and spiral trajectories are used only for
evaluation. The ellipse trajectories used
during training expose the interceptor to a wide range of maneuvers to
intercept the intruder. The lemniscate and spiral trajectories
evaluate the generalizability of the approach to more complex
motions. Each trajectory is generated by randomly sampling
parameters (see~\cref{appx-ssec:intru-traj}), including a constant speed for the
intruder to follow and geometric parameters such as the major
axis, aspect ratio, orientation, and ascent rate.

\textbf{Implementation Details:}
During both training and evaluation, rollouts that terminate early
due to either successful interception or exceeding environment bounds,
are frozen until all parallel rollouts complete execution to
synchronously update the policy parameters. Experiments were conducted
on an NVIDIA RTX 5090 GPU. Unless otherwise specified, policies were
trained for 1500 updates using a horizon length of 64 steps across 512
parallel environments, corresponding to approximately 49.15 million
steps.  We evaluate the learnt policies in 110,000 environments.

\begin{figure}
    \centering
    \subfloat[Success rate during training]{\label{fig:algo-train-success}
        \includegraphics[width=0.49\linewidth]{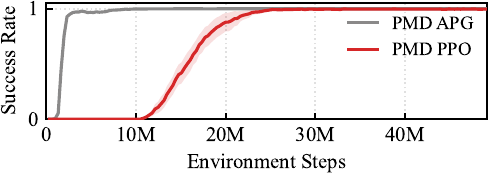}%
    }\hfill
    \subfloat[Episode length during training]{\label{fig:algo-train-eplen}
        \includegraphics[width=0.49\linewidth]{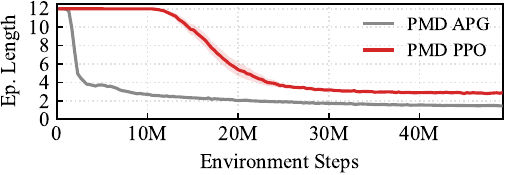}%
    }
    \caption{\label{fig:algo-train-curves}Training success rate and episode
    length variation with environment steps demonstrate that APG is more
    sample-efficient than PPO.}
    \centering
    \subfloat[Ellipse]{\label{fig:eval-ellipse}
        \includegraphics[width=0.32\linewidth]{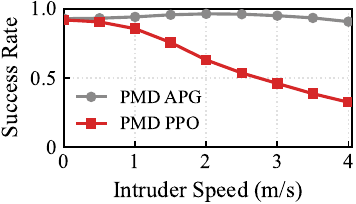}%
    }\hfill
    \subfloat[Spiral]{\label{fig:eval-spiral}
        \includegraphics[width=0.32\linewidth]{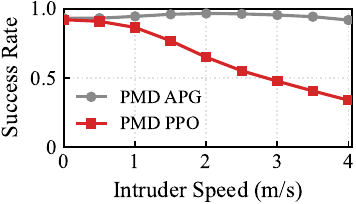}%
    }\hfill
    \subfloat[Lemniscate]{\label{fig:eval-lemniscate}
        \includegraphics[width=0.32\linewidth]{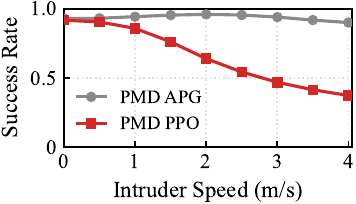}%
    }
    \caption{\label{fig:algo-predef-eval}
      The success rates obtained while varying intruder speeds during
      evaluation demonstrate that using APG enables a higher interception
      accuracy than PPO for all three types of intruder trajectories.}
\end{figure}

\textbf{Evaluation of APG and PPO using Point Mass Dynamics (PMD):}
Prior works~\cite{zhang2025learning,lee2026quadrotor} have leveraged
simplified point mass dynamics and differentiable physics for
quadrotor navigation tasks due to their computational
efficiency. Given this prior work, we conduct an analysis to evaluate
the performance of APG using a simplified point mass dynamics model
and compare its performance against the state-of-the-art model-free
baseline PPO~\citep{schulman2017proximal}. Consistent with prior
work, we employ domain randomization (see~\cref{tab:hp-dyn-pmd}) to help
the PMD policies generalize to the quadrotor dynamics.

The baseline PPO policy is implemented using GAE \cite{schulman2015high} and two
separate networks for the actor and critic. The actor and critic networks
leverage the same architecture used in our proposed method (the parameters used
for PPO training may be found in \cref{tab:hp-algo}). For a fair
comparison, both policies share the same state and action space, and are trained
against the same objective function in~\cref{eq:objective_fn}. The
interceptor is initialized in a stable hover state at the beginning of each
episode.

To analyze the performance, we generate two sets of plots. First, we provide
a mean and standard error plot (\cref{fig:algo-train-curves}) for success
rate and episode length observed during training for both approaches
based on $10$ training runs with different random number generator seeds.
The second set of plots (\cref{fig:algo-predef-eval}) evaluates the
generalizability of each approach across different trajectories and velocity
profiles of the intruder. We find that APG is more sample-efficient than PPO
(\cref{fig:algo-train-success}), which is consistent with prior
work~\citep{zhang2025learning}.~\Cref{fig:algo-train-eplen} shows that PPO takes
longer to intercept the intruder. \Cref{fig:algo-predef-eval} shows the success rates for the ellipse, spiral, and lemniscate trajectories. The PMD APG approach consistently
outperforms the PMD PPO approach and widens the success gap as the speed of the
intruder is increased.

These analyses assume a relatively large platform (i.e., \SI{2.65}{\kilo\gram})
with a thrust-to-weight ratio of $2.7$, which affects the agility of the
platform and causes motor saturation at high intruder speeds beyond
\SI{4}{\meter\per\second}. The PMD and Quad interceptor platform parameters are
provided in~\cref{tab:hp-dyn-pmd} and~\cref{tab:hp-dyn-quad}, respectively.

\begin{figure}
    \centering

    \subfloat[Success rate during training]{\label{fig:dyn-train-success}
        \includegraphics[width=0.49\linewidth]{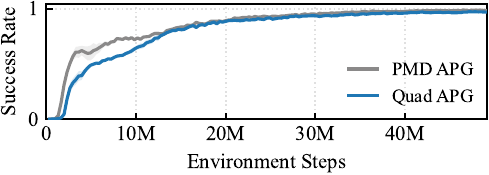}%
    }\hfill
    \subfloat[Episode length during training]{\label{fig:dyn-train-eplen}
        \includegraphics[width=0.49\linewidth]{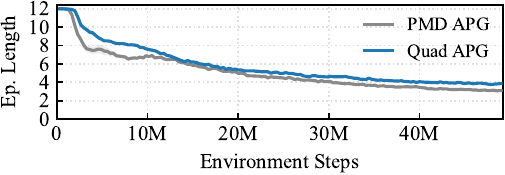}%
    }
    \caption{\label{fig:dyn-train-curves}
      Training success rate and episode length variation with environment
      steps show a similar rate of convergence when simplified point mass
      dynamics and nonlinear quadrotor dynamics models are used with APG.}
    \centering
    \subfloat[Ellipse]{\label{fig:dyn-eval-ellipse}
        \includegraphics[width=0.32\linewidth]{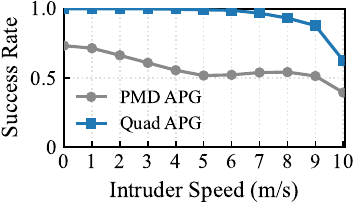}%
    }\hfill
    \subfloat[Spiral]{\label{fig:dyn-eval-spiral}
        \includegraphics[width=0.32\linewidth]{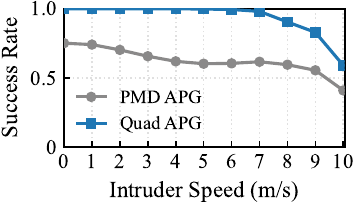}%
    }\hfill
    \subfloat[Lemniscate]{\label{fig:dyn-eval-lemniscate}
        \includegraphics[width=0.32\linewidth]{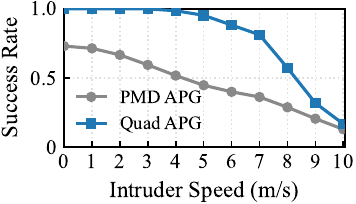}%
    }
    \caption{\label{fig:dyn-predef-eval}
      The success rates obtained while varying intruder speeds during
      evaluation demonstrate that using the nonlinear quadrotor dynamics
      model enables a higher interception accuracy
      compared to using simplified point mass dynamics for all three
      types of intruder trajectories.}
\end{figure}

\textbf{Evaluation of PMD APG Compared to APG Trained with Quadrotor
Dynamics:} While prior works~\cite{zhang2025learning,lee2026quadrotor}
have shown that simplified point mass
dynamics are sufficient for quadrotor navigation, the navigation
problem seeks to achieve a single setpoint. The interception
problem is fundamentally different as it requires more agile maneuvers
to catch a high-speed moving intruder. For this reason, we
opt for using the nonlinear quadrotor dynamics model.

For these experiments, we reduce the interceptor mass to
\SI{1}{\kilo\gram} effectively increasing the thrust-to-weight ratio
and platform agility of the interceptor. In the next section, we
provide results to quantify the performance for a larger and higher
range of intruder speeds with a more agile interceptor. Given the
superior performance of PMD APG compared to PMD PPO, we compare PMD
APG against APG trained using quadrotor dynamics (Quad APG).

\begin{figure}
    \centering
    \subfloat{
        \includegraphics[trim=10 0 10 30,clip,width=0.32\linewidth]{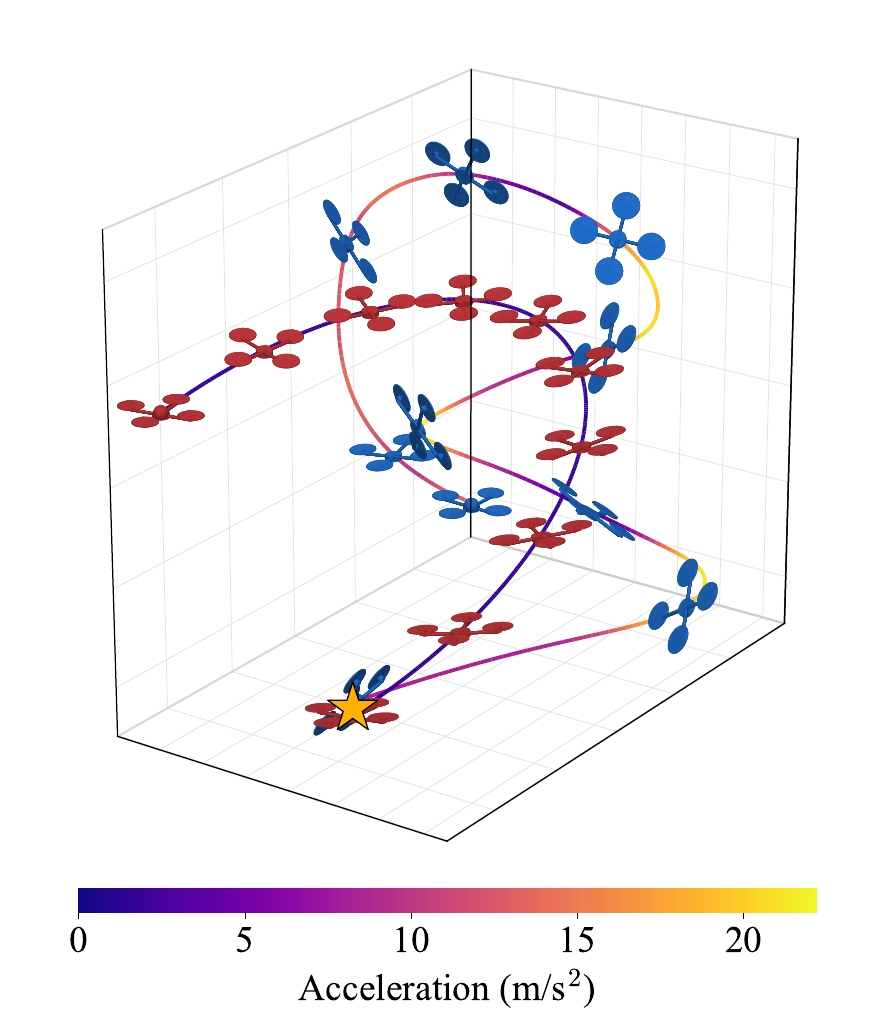}%
    }\hfill
    \subfloat{
        \includegraphics[trim=10 0 10 30,clip,width=0.32\linewidth]{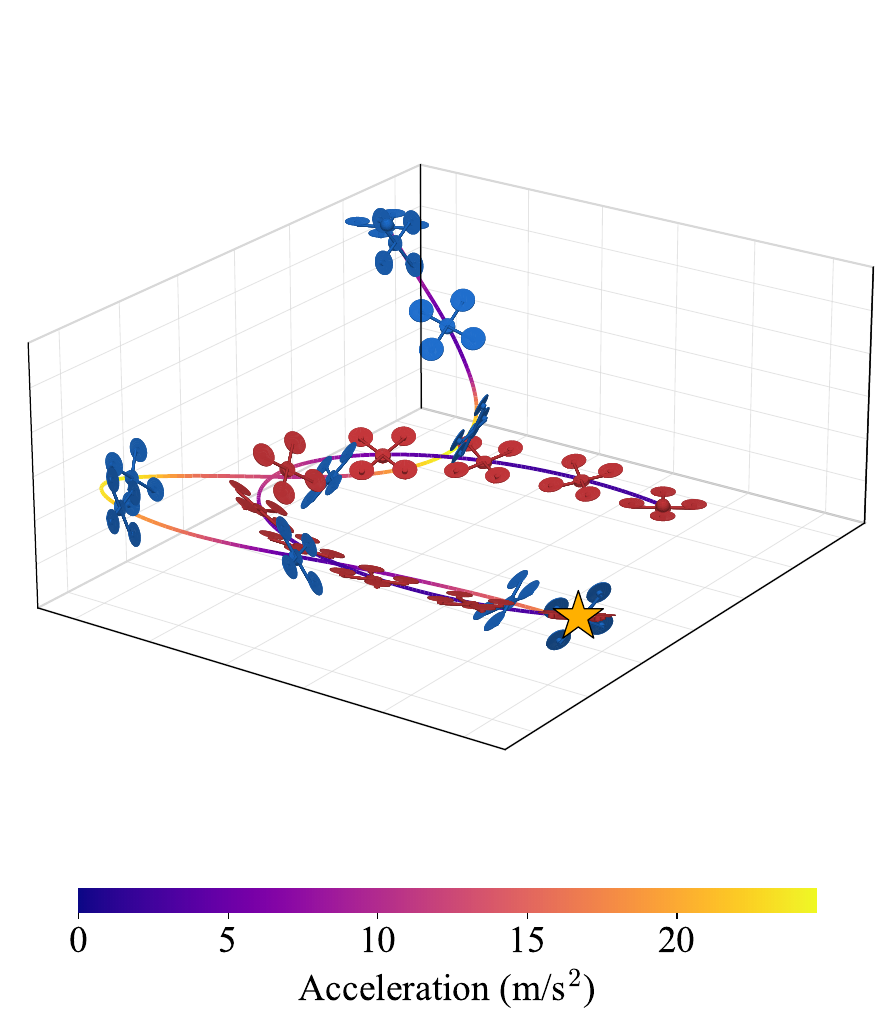}%
    }\hfill
    \subfloat{
        \includegraphics[trim=10 0 10 30,clip,width=0.32\linewidth]{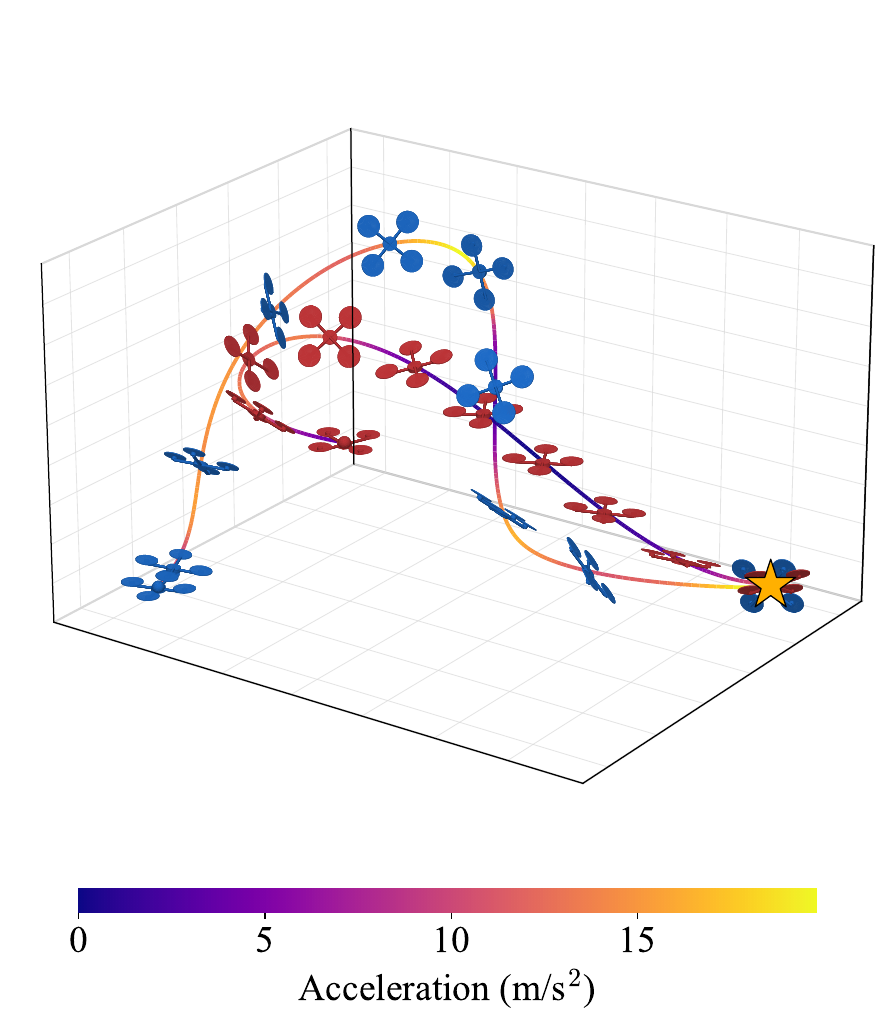}%
    }\\\setcounter{subfigure}{0}
    \subfloat[Ellipse]{\label{fig:ellipse-rollout}
        \includegraphics[trim=10 0 0 75,clip,width=0.32\linewidth]{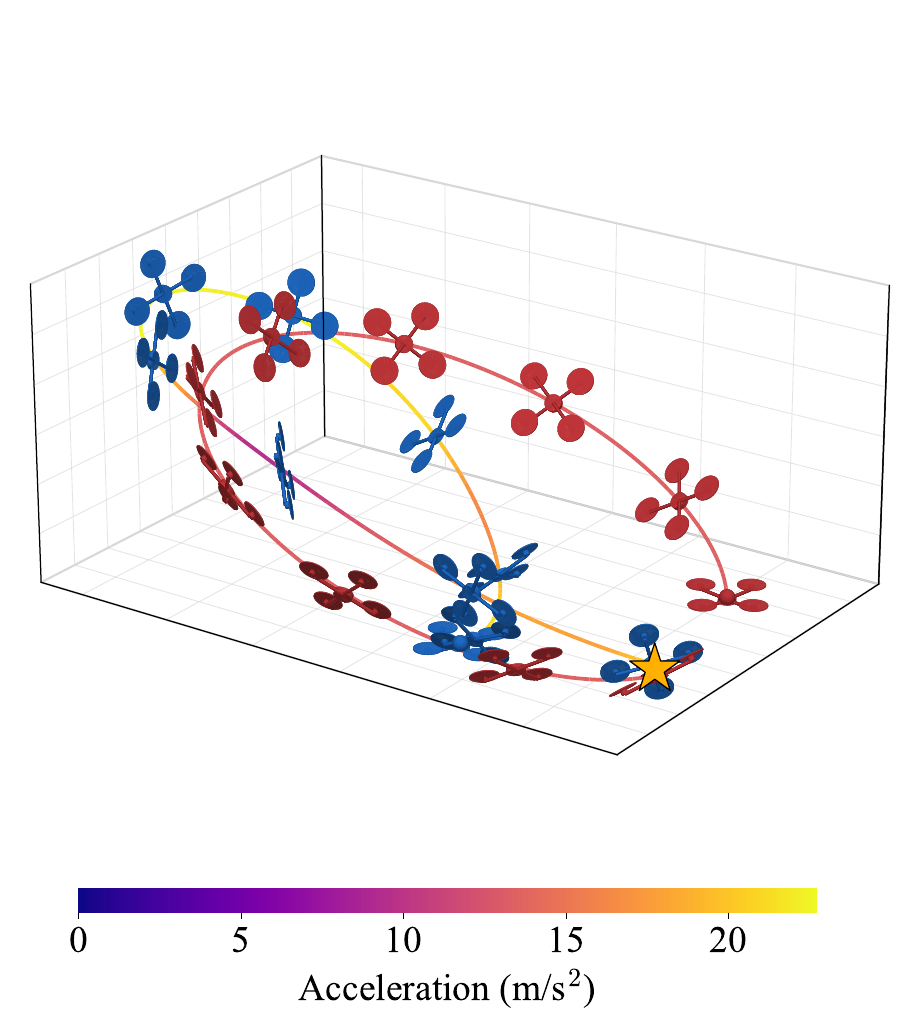}%
    }\hfill
    \subfloat[Spiral]{\label{fig:spiral-rollout}
        \includegraphics[trim=10 0 0 75,clip,width=0.32\linewidth]{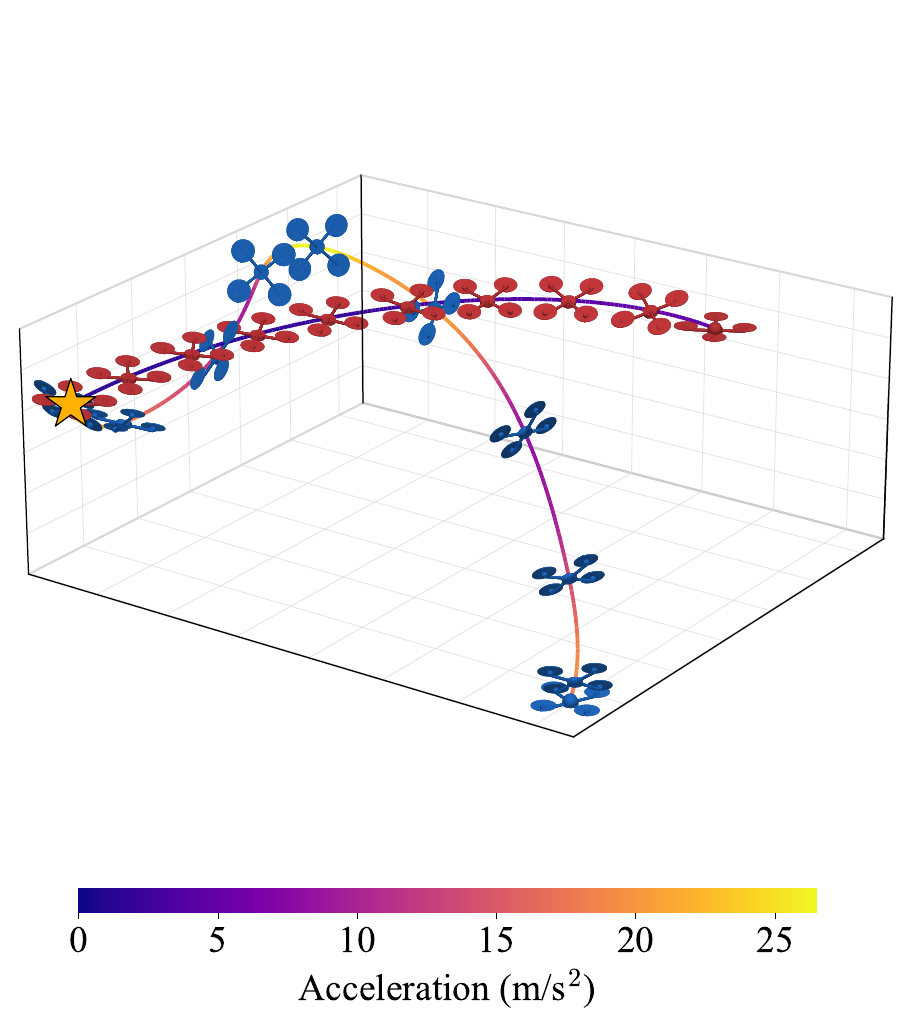}%
    }\hfill
    \subfloat[Lemniscate]{\label{fig:lemniscate-rollout}
        \includegraphics[trim=10 0 0 75,clip,width=0.32\linewidth]{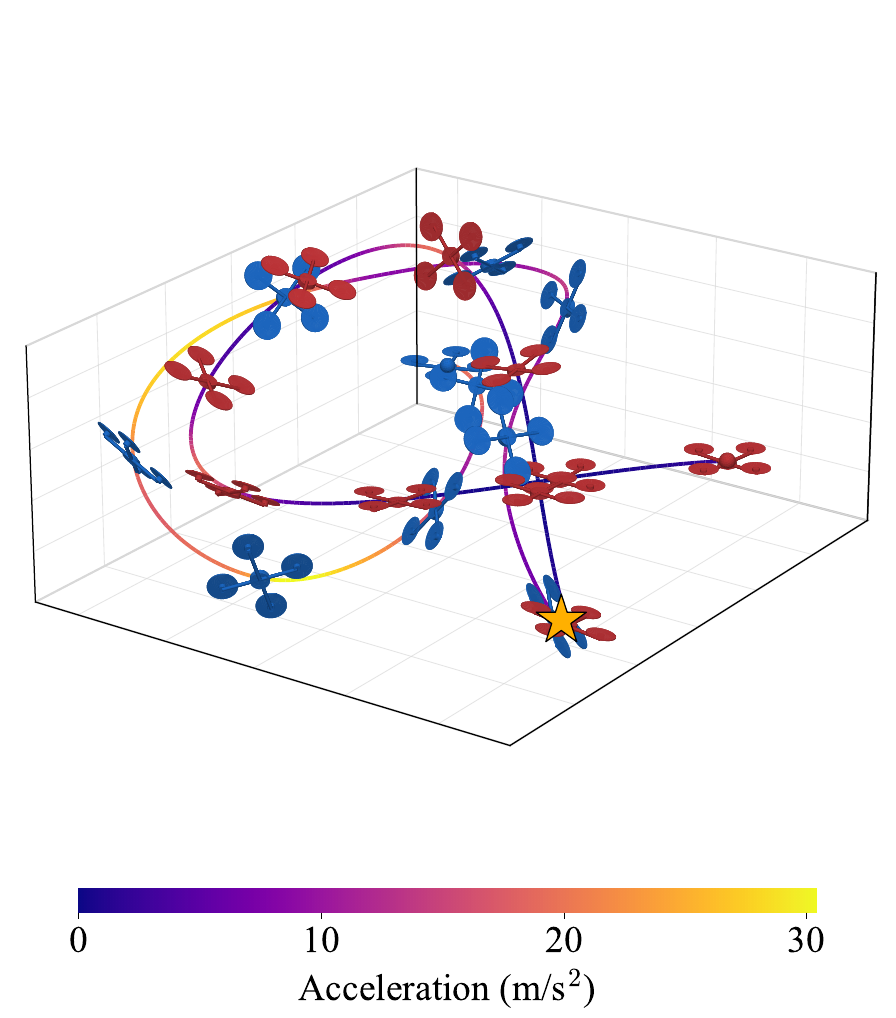}%
    }
    \caption{\label{fig:viz-rollouts}Example rollouts with
      acceleration heatmaps of the proposed Quad APG policy. Collision
      between the intruder (red) and the interceptor (blue) occurs at the
      point marked with a star.}
\end{figure}

To evaluate the performance of PMD APG and Quad APG, we provide two
sets of plots. The first set evaluates the training performance
(see~\cref{fig:dyn-train-curves}). We observe PMD and Quad dynamics
models require a similar number of environment steps for convergence.
During evaluation, which is shown
in~\cref{fig:dyn-predef-eval}, we see that the success rate for the
Quad APG policy is substantially higher compared to the PMD APG policy.
The Quad APG policy is able to intercept approximately 60\% of intruders at speeds up to
\SI{8}{\meter\per\second} (compared to approximately 30\% of intruders
for PMD APG). For each of the ellipse, spiral, and lemniscate
evaluation trajectories, the Quad APG policy outperforms the PMD APG
policy on average by 37\%, 33\%, and 31\%, respectively.~\Cref{fig:viz-rollouts}
provides qualitative examples of rollouts for the Quad APG policy using the three trajectory
types.

\textbf{Training Times:} The Quad APG policy takes approximately $137$ minutes,
the PMD APG policy takes $10$ minutes, and PMD PPO takes approximately 7 minutes. 
Though significantly faster to train, the low fidelity of the
point mass model means it cannot capture the full range of rigid-body
dynamics that a true quadrotor model exhibits.

\section{Conclusion}
This paper presented a methodology to intercept an intruder by
learning agile maneuvers using differentiable simulation.
The interceptor state and 3D unit
direction vector to the intruder are provided as inputs to the policy
to infer the high-level control commands for interception.
The policy parameters are optimized using the sample-efficient
analytical policy gradients.
Training through the quadrotor dynamics instead of
simplified point mass dynamics yields higher interception rates ($\geq$ 30\%
on average) for a wide range of intruder speeds
(up to \SI{10}{\meter\per\second}).

\textbf{Limitations:} The current approach assumes that the intruder
uses a constant speed to execute the trajectory; however, this is not
representative of an adversarial scenario. Future work includes
developing a policy that intercepts adversarial evaders.  Our approach
also assumes that the intruder is always within the field of view of
the interceptor and does not incorporate sensor or detection
uncertainty. Hardware validation also remains an area of future work.


\acknowledgments{This research was supported in part by an AI2C Seed
grant and the NVIDIA Academic Grant Program.}


\bibliography{refs}  

\clearpage
\appendix
\crefname{appendix}{appendix}{appendices}
\Crefname{appendix}{Appendix}{Appendices}

\section{Appendix}
\subsection{Parametric Intruder Trajectories\label[appendix]{appx-ssec:intru-traj}}
\begin{table}[H]
  \centering
  \caption{Intruder trajectory families. $\axisrad$~=~semi-axis, $\aratio$~=~axis
    ratio, $\zrate$~=~spiral climb per radian.}
  \label{tab:trajectory_families}
  \begin{tabular}{@{}llll@{}}
    \toprule
    Family & In-plane $\vh(s)$ & Vertical $z(s)$ & Tilt $\Rtilt$ \\
    \midrule
    Ellipse    & $(\axisrad \cos s, \axisrad\aratio\sin s, 0)$        & $0$                   & per-env $(\phi,\theta)$ \\
    Spiral     & $(\axisrad \cos s, \axisrad\aratio\sin s, 0)$        & $\zrate s$ & identity (flat baseline) \\
    Lemniscate & $(\axisrad \cos s, \tfrac{\axisrad\aratio}{2}\sin 2s, 0)$ & $0$              & per-env $(\phi,\theta)$ \\
    \bottomrule
  \end{tabular}
\end{table}

Each intruder trajectory is a 3D curve parameterized by a scalar phase $s$,
\begin{equation}
  \vp(s) = \mathbf{c}_0 + \Rtilt \vh(s) + z(s)\zhat ,
  \label{eq:curve_model}
\end{equation}
where $\mathbf{c}_0$ is the curve center, $\vh(s)$ is an in-plane curve in the
$xy$-plane, and $z(s)$
is a vertical component along the world $\zhat$ axis.

We instantiate three families of trajectories: ellipse, spiral, and
lemniscate (\cref{tab:trajectory_families}). The tilt $\Rtilt$ is a
per-rollout $\mathrm{ZYX}$-Euler rotation $\Rtilt = R_z(\psi)R_y(\theta)R_x(\phi)$ that
lifts the planar curve into 3D. During training, we use constant speed 3D elliptical
trajectories without a dynamic feasibility check. For evaluation, we use all
three trajectory types but with an upper bound on maximum acceleration.
In the following, $\dot{()}$ and $()'$ denote derivatives with respect to time
and phase, respectively.

\subsubsection{Training}
Let the desired intruder speed be $\vtgt$. We ensure that the elliptical
realization of the parametric curve in~\cref{eq:curve_model} follows this target
speed by setting $\dot{s} = \vtgt / \norm{\vc(s)}$ in $\dot{\vp} = \vc(s)
\dot{s}$, where $\vc(s) = \Rtilt \vh'(s) + z'(s)\zhat$. The sign of $\vtgt$
determines the intruder's direction of traversing the curve.

\subsubsection{Evaluation}
To impose an upper bound on acceleration, we derive a feasibility predicate
directly from the quadrotor model so the evaluation set of trajectories is
realizable. The quadrotor produces net world-frame acceleration
\begin{equation}
  \vanet = \frac{T_c}{\mass}\vb - \grav \zhat,
  \label{eq:quad_world_acceleration}
\end{equation}
where $\vb$ is the body thrust axis and $T_c$ the collective thrust. To
realize a demanded acceleration $\vad$, the rotors must produce the thrust
acceleration $\vatcmd = \vad + \grav \zhat$. This is feasible if
\begin{equation}
  \norm{\vatcmd} \le A_{\max}
  \qquad\text{and}\qquad
  a_{t,z}^{\mathrm{cmd}} \ge 0,
  \label{eq:quad_feasibility_condition}
\end{equation}
where $A_{\max}$ is a maximum acceleration chosen using the maximum RPM values of the four rotors ($N_m = 4$)
and a degradation factor $\eta$ to account for motor time constant, control bandwidth, and error-correction margin.
Formally,
\begin{equation}
  A_{\max} = \eta \frac{N_m F_m}{\mass}
  \label{eq:quad_total_thrust_budget}
\end{equation}
such that $F_m = f_2 \rpmmax^2 + f_1 \rpmmax + f_0$ is the maximum available per-rotor force and
$\eta = 0.6$.

A trajectory represented by the family in~\cref{eq:curve_model} is feasible if
every point on the curve satisfies~\cref{eq:quad_feasibility_condition}.
For the elliptical and lemniscate trajectories, we use a dense grid of phase values $s$ and ensure
\begin{equation}
  \max_s \norm{\va(s) + \grav\,\zhat} \le A_{\max}
  \quad\text{and}\quad
  \min_s a_z(s) \ge -\beta\,\grav,
  \label{eq:ell-lemi-constraint}
\end{equation}
where the downward-acceleration fraction $\beta = 0.6$ leaves margin so
the attitude controller does not saturate on the descending arc of a
tilted orbit. The expression
for $\va(s)$ can be derived from~\cref{eq:curve_model} as
\begin{equation}
  \va(s) = \vtgt^2\left[\frac{\vc'(s)}{\norm{\vc(s)}^2}
      - \vc(s)\frac{\vc(s)\cdot\vc'(s)}{\norm{\vc(s)}^4}\right].
  \label{eq:target_trajectory_acceleration}
\end{equation}
The value $a_z(s)$ in~\cref{eq:ell-lemi-constraint} is the $z$-component
of~\cref{eq:target_trajectory_acceleration}. On the other hand, for the spiral
trajectory, we impose $\vtgt^2\,\kappa_{\max} \le \mass\grav$, where $\kappa_{\max} =
\max_s \kappa(s)$ is the peak in-plane curvature, which is derived
from~\cref{eq:curve_model} as $\kappa(s) =
\norm{\vc(s)\times\vc'(s)} / \norm{\vc(s)}^3$.

We use the following algorithm to generate a desired number of these trajectories for different speed magnitudes.
\begin{algorithm}[ht!]
  \caption{Generating a pool of feasible trajectories for a range of speed magnitudes}
  \label{alg:gen}
  \begin{algorithmic}[1]
    \Require speed magnitudes, desired number of trajectories, shape-parameter
    ranges, feasibility test, pool size, draw budget
    \For{each speed magnitude}
      \State \textit{feasible pool} $\gets$ empty
      \State \textit{draws} $\gets 0$
      \While{\textit{feasible pool} $<$ pool size \textbf{and}
             \textit{draws} $<$ draw budget}
        \State draw shape parameters uniformly from the ranges
          \Comment{semi-axis, axis ratio, tilt, $\zrate$}
        \State \textit{draws} $\gets \textit{draws} + 1$
        \If{the shape is dynamically feasible at this speed}
            \Comment{feasibility test}
          \State add the shape to the \textit{feasible pool}
        \EndIf
      \EndWhile
      \If{size(\textit{feasible pool}) $<$ desired number of trajectories}
        \State \textbf{abort}: feasible region too thin
          \Comment{widen the ranges or lower the top speed}
      \EndIf
      \State \textit{kept} $\gets$ farthest-point selection from
        the \textit{feasible pool}
        \Comment{max-min in normalized parameter space}
      \State signed-split: assign traversal direction
      \State output each \textit{kept} shape and its signed speed
    \EndFor
  \end{algorithmic}
\end{algorithm}

The algorithm has three stages. First,~\emph{rejection sampling} draws shape
parameters uniformly and keeps those that pass the feasibility
test, building a pool up to a specified pool size. Second,~\emph{farthest-point
selection} greedily selects the requested number of pool members that maximize
the minimum pairwise distance in normalized parameter space, so the \emph{kept}
set is geometrically diverse rather than clumped by feasible-region density.
Third,~\emph{signed split} assigns a positive direction to half and a negative
direction to the other half (reverse traversal) pool of trajectories. Each
generation run reports acceptance rate and diversity diagnostics (e.g., closest-pair
spread, per-axis coverage, etc.) and aborts if any (family, speed) bucket cannot be
filled, so a thin feasible region cannot silently degrade the evaluation
trajectory set.

\subsection{Hyperparameter Sweeping\label{app:paramsweep}}

\subsubsection{Loss Coefficients}
For the loss coefficients, $\lambda_{\mathrm{align}}, \lambda_{\mathrm{close}},
\lambda_{\mathrm{acc}}, \lambda_{\mathrm{jerk}}, \lambda_{\mathrm{vmax}}, \text{ and }
\lambda_{\mathrm{yaw}}$ in~\cref{eq:objective_fn}, we set up hyperparameter
sweeping as follows.

\textbf{Scale:} The objective is nearly invariant to a global rescaling
$\bm{\lambda}\!\to\!\alpha\bm{\lambda}$ absorbed into the policy step size: only
the ratios of the weights matter. We anchor $
\lambda_{\mathrm{align}}\mathcal{L}=5$ and search for the remaining weights and
the learning rate relative to the anchor.

\textbf{Trial Objectives:} Each trial trains a policy from scratch on the point mass
dynamics under a candidate $\bm{\lambda}$, then evaluates it on the
high-fidelity quadrotor model. This weight vector is used
for all training runs in this paper. The objective is to minimize the scalar
\begin{equation}
  J = 300(1 - \mathrm{SR}) + L_{p99},
  \label{eq:obj}
\end{equation}
where $\mathrm{SR}\in[0,1]$ is the interception success rate over the evaluation
episodes and $L_{p99}$ is the 99th-percentile episode length
(time-to-intercept).  Interceptions that take more than 12 seconds are
considered failures.  The main reason for introducing time-to-intercept into the
objective is to break ties between policies with similar success rates, which is
common in the high-success regime.  The large weight on the success rate ensures
that the optimizer prioritizes improving the success rate until it is close to
100\%, at which point it will focus on reducing time-to-intercept.  This design
encourages the discovery of policies that not only succeed but also do so
efficiently, rather than optimizing for speed at the expense of reliability.
Specifically, 99th-percentile time-to-intercept is used instead of the mean or
median to focus on improving the worst-case performance, which is critical for
real-world applications where consistent performance is often more important
than average performance.

\textbf{Search procedure:} We use Optuna's TPE sampler via
Hydra's Optuna sweeper plugin\footnote{\url{https://hydra.cc/docs/plugins/optuna_sweeper/}}.
We run it in multivariate
mode with an initial $16$ random startup trials before the model engages. We run a total of
$64$ trials per study. All positive weights and the learning rate are given
log-uniform priors (they span orders of magnitude); bounded ratios such as the
learning-rate decay are also log-uniform.

\textbf{Iterative range refinement:} We first manually adjust the
hyperparameters to a suboptimal set that achieves a baseline level of
performance showing convergence.  We then run a sweep over a wide range around
the empirically tested weight sets.  However, many trials will be spent in
regions of the weight space that are suboptimal or even detrimental to learning.
To address this, we run a sequence of studies, after each of which we re-compute
the weight set.  After every study we compute, per weight: (i) FANOVA and
mean-decrease-impurity, (ii) the Spearman rank correlation between the
log-weight and $J$ which is a monotonicity measure, (iii) a summary of the
top-$10\%$ trials, and (iv) the 1D slice plot of $J$ against each weight. Using
the shape of the 1D slice, we adjust the next weight set.

\begin{figure}
    \centering
    \subfloat{
        \includegraphics[width=0.32\linewidth]{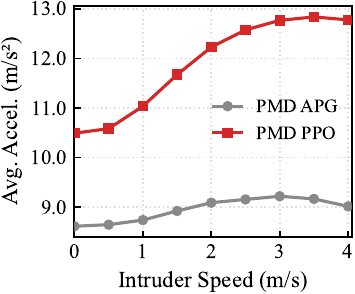}%
    }\hfill
    \subfloat{
        \includegraphics[width=0.32\linewidth]{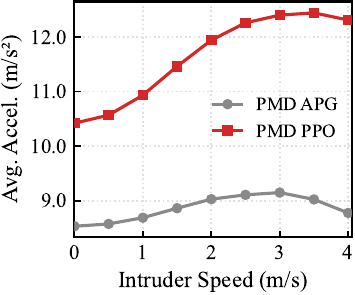}%
    }\hfill
    \subfloat{
        \includegraphics[width=0.32\linewidth]{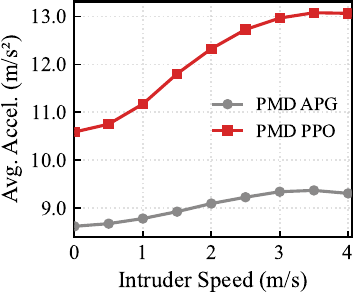}%
    }
    \setcounter{subfigure}{0}
    \subfloat[Ellipse]{
        \includegraphics[width=0.32\linewidth]{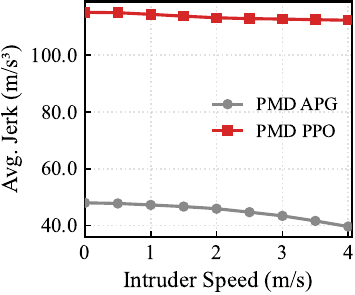}%
    }\hfill
    \subfloat[Lemniscate]{
        \includegraphics[width=0.32\linewidth]{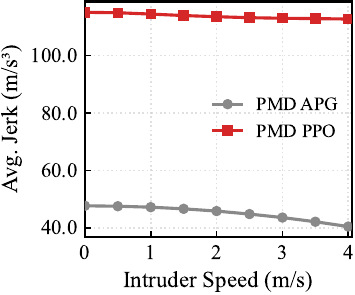}%
    }\hfill
    \subfloat[Spiral]{
        \includegraphics[width=0.32\linewidth]{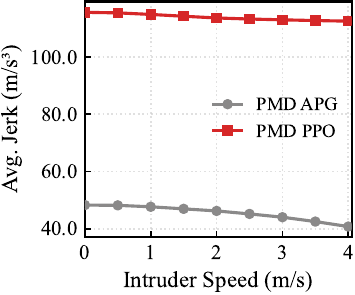}%
    }
    \caption{Average acceleration (top row) and jerk (bottom row) across
    intruder speeds for successful rollouts using APG and PPO. We observe that
    using the PPO-trained policy leads to higher acceleration and jerk for the
    interceptor compared to APG.}
    \label{fig:acc-jerk-algo}
\end{figure}%
\begin{figure}
    \centering
    \subfloat{
        \includegraphics[width=0.32\linewidth]{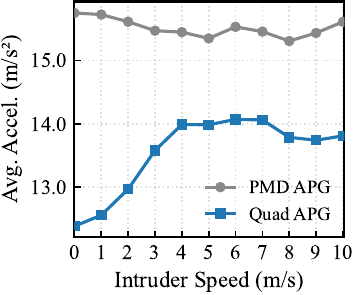}%
    }\hfill
    \subfloat{
        \includegraphics[width=0.32\linewidth]{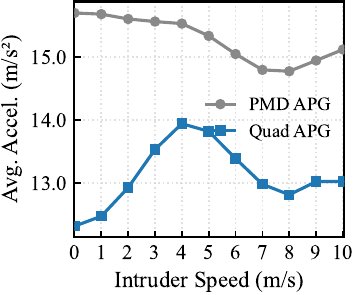}%
    }\hfill
    \subfloat{
        \includegraphics[width=0.32\linewidth]{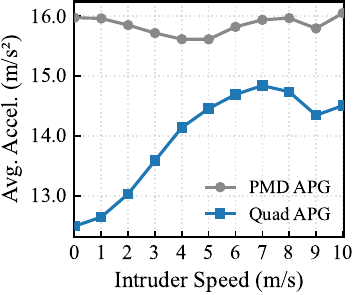}%
    }
    \setcounter{subfigure}{0}
    \subfloat[Ellipse]{
        \includegraphics[width=0.32\linewidth]{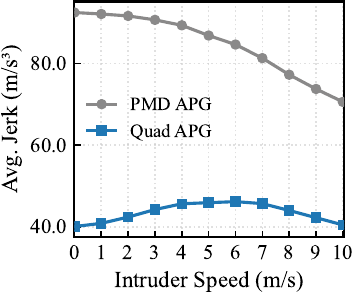}%
    }\hfill
    \subfloat[Lemniscate]{
        \includegraphics[width=0.32\linewidth]{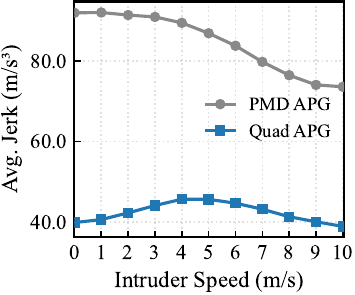}%
    }\hfill
    \subfloat[Spiral]{
        \includegraphics[width=0.32\linewidth]{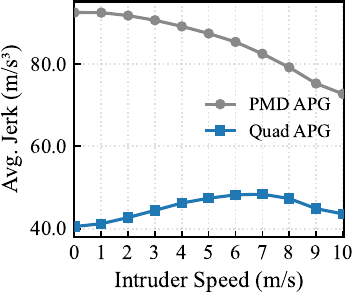}%
    }
    \caption{Average acceleration (top row) and jerk (bottom row) across
    intruder speeds for successful rollouts using the simplified point mass
    (PMD) and nonlinear quadrotor dynamics (Quad) models with APG. Leveraging the Quad
    dynamics model provides interception with lower control effort than the PMD
    case.}
    \label{fig:acc-jerk-dyn}
\end{figure}

\subsection{Additional Performance Analysis}
In addition to the interception success rate and episode
length, we characterize the behavior of the policies by analyzing their
acceleration and jerk profiles. We calculated these values for:
algorithm comparison (APG vs.\ PPO) and training dynamics (full quadrotor vs.\
PMD), and report per-step acceleration and jerk conditioned on the intruder speed.

Policies that produce large acceleration commands tend to behave more
aggressively when pursuing the intruder. While aggressive acceleration commands
can sometimes help with interception performance, they may also lead to
catastrophic overshooting when tracking agile trajectories.  Overshooting can
leave the policy in undesirable states that are difficult to recover from.
Recovering from these scenarios often requires abrupt corrective actions,
resulting in increased jerk and higher overall control effort. In our analysis,
we study how the acceleration and jerk characteristics learned by a policy
correlate with interception success across different trajectory classes.

We first compare the acceleration and jerk profiles of APG and PPO, both trained
using simplified point mass dynamics and evaluated on the nonlinear quadrotor dynamics
model (see~\cref{fig:acc-jerk-algo}). Similarly, to further
understand the performance differences between APG policies trained with
these dynamics models, we compare their acceleration
and jerk profiles in \cref{fig:acc-jerk-dyn}. For both analyses, we report
results only for episodes that terminate with a successful interception of the
intruder. In both cases we observe a lower control effort required by the interceptor
when APG and nonlinear quadrotor dynamics are used for training.

\subsection{Parameters\label{app:final_params}}
\begin{table}
\centering
\caption{Shared across both benchmarks: training setup (left) and loss-signal
weights (right).}
\label{tab:hp-shared}
\begin{minipage}[t]{0.46\linewidth}
\centering
\begin{tabular}{lr}
\toprule
\textbf{Training setup} & \textbf{Value} \\
\midrule
Environments $[\nenv]$              & 512 \\
Updates $[\nupd]$                   & 1500 \\
Max episode steps $[\maxsteps]$     & 600 \\
Seeds $[\nseed]$                    & 10 \\
Capture radius $[\captureradius]$ (m) & 0.3 \\
Escape radius $[\escaperadius]$ (m) & 100.0 \\
Spawning box size $[\sidelength]$ (m) & 10.0 \\
\bottomrule
\end{tabular}
\end{minipage}\hfill
\begin{minipage}[t]{0.46\linewidth}
\centering
\begin{tabular}{lr}
\toprule
\textbf{Signal weight} & \textbf{Value} \\
\midrule
$\lamvalign$    & 5.0 \\
$\lamvclose$    & 1.07 \\
$\lamacc$       & 0.0015 \\
$\lamjerk$      & $1.96\times10^{-4}$ \\
$\lamvmax$      & 0.4 \\
$\lamyaw$       & 0.064 \\
\bottomrule
\end{tabular}
\end{minipage}
\end{table}

\begin{table}
\centering
\caption{Algorithm hyperparameters. Dashes indicate the entry is not applicable
to that algorithm.}
\label{tab:hp-algo}
\begin{tabular}{lrr}
\toprule
\textbf{Parameter}                          & \textbf{APG} & \textbf{PPO} \\
\midrule
Horizon $[\horizon]$                        & 64    & 64 \\
Number of environments $[\nenv]$            & 512   & 512 \\
Actor learning rate $[\lrA]$                & $1.64\times10^{-3}$ & $1.51\times10^{-3}$ \\
Final learning rate $[\lrfinal]$            & $5\times10^{-4}$    & --- \\
LR decay ratio $[\lrdecay]$                 & 0.186 & --- \\
Weight decay $[\wdecay]$                    & 0.01  & --- \\
Max gradient norm $[\gmax]$                 & 5.0   & 1/1 \\
Discount $[\discount]$                      & 1     & 0.99 \\
Critic learning rate $[\lrC]$               & ---   & $1\times10^{-3}$ \\
Epochs per update $[\Kepoch]$               & ---   & 1 \\
Minibatches $[\Nmb]$                        & ---   & 8 \\
GAE $[\gaelam]$                             & ---   & 0.95 \\
PPO clip $[\ppoclip]$                       & ---   & 0.2 \\
Value loss weight $[\cval]$                 & ---   & 0.5 \\
Entropy weight $[\cent]$                    & ---   & 0.01 \\
\bottomrule
\end{tabular}
\end{table}

The experimental configuration is summarized
in~\cref{tab:hp-shared,tab:hp-algo,tab:hp-traj,tab:hp-dyn-pmd,tab:hp-dyn-quad}.

\Cref{tab:hp-shared} reports the settings used for both the algorithm
benchmark (PMD PPO vs.\ PMD APG) and the dynamics benchmark (PMD APG vs.\ Quad
APG). In all experiments, $\nenv = 512$ parallel environments were trained for
$\nupd = 1500$ updates, with episode capped at $\maxsteps = 600$ steps. Results
were averaged across $\nseed = 10$ unique random seeds. For each seed, the
policy was retrained from scratch. We set the control timestep and maximum
episode length to be $\dtctrl = 0.02$~s and $\maxsteps = 600$ respectively
(total episode duration is $12$~s). These parameters were selected to provide
sufficient time for interception under the present configuration while keeping
episodes short enough for efficient training and stable policy optimization. The
capture radius was set to $\captureradius = 0.3$~m, corresponding approximately
to the physical extent of a large intruder vehicle.

Observations and actions are expressed in the start frame, defined by the
agent orientation at initialization rather than the global world frame.
Anchoring commands to this frame removes dependence on a fixed global coordinate
frame, allowing the learned policy to behave consistently regardless of the
interceptor's initial orientation.

\Cref{tab:hp-algo} lists the algorithm-specific hyperparameters for APG and PPO
side by side, with dashes indicating entries that are not applicable to a given
method.

\Cref{tab:hp-traj} specifies the intruder trajectory randomization
applied at the start of each episode. Each parameter is sampled uniformly over
the range shown in the table.

Finally,~\cref{tab:hp-dyn-pmd,tab:hp-dyn-quad} report the parameters for the two
dynamics models used throughout the experiments. For the point mass dynamics
(PMD) model, we identify the subset of parameters subjected to domain
randomization during training to improve transferability to nonlinear quadrotor
dynamics evaluation. The nonlinear
quadrotor dynamics model parameters include the low-level controller parameters
used to track policy outputs. For the algorithm benchmark, the modeled quadrotor
mass during evaluation was set to $2.65$~kg. For the APG dynamics comparison,
the mass was reduced to $1.0$~kg to increase the thrust-to-weight ratio, thereby
simulating a more agile interceptor capable of handling higher-speed intruders.

\begin{table}
\centering
\caption{Intruder trajectory randomization during training. The speed range depends on benchmark:
$\mathcal{U}(-4,4)$~m/s for \autoref{fig:algo-predef-eval} and
$\mathcal{U}(-10,10)$~m/s for \autoref{fig:dyn-predef-eval}.}
\label{tab:hp-traj}
\begin{tabular}{lr}
\toprule
\textbf{Parameter}              & \textbf{Value} \\
\midrule
Speed $[\vtgt]$ (m/s)           & $\mathcal{U}(-10,10)$ / $\mathcal{U}(-4,4)$ \\
Tilt rpy $[\tiltrpy]$ (rad)     & $\mathcal{U}(-0.5,\,0.5)$ \\
Axis radius $[\axisrad]$ (m)    & $\mathcal{U}(4,\,8)$ \\
Aspect ratio $[\aratio]$        & $\mathcal{U}(0.5,\,1.5)$ \\
Curve type                      & ellipse \\
\bottomrule
\end{tabular}
\end{table}

\begin{table}
\centering
\caption{Point mass dynamics parameters with
domain-randomization ranges.}
\label{tab:hp-dyn-pmd}
\centering
\begin{tabular}{lr}
\toprule
\textbf{PMD} & \textbf{Value} \\
\midrule
Sim.\ timestep $[\dtsim]$ (s)            & 0.0025 \\
Control timestep $[\dtctrl]$ (s)         & 0.02 \\
Gravity $[\grav]$ (m/s$^{2}$)            & 9.807 \\
Quadratic drag $[\kdone]$                & 0.1 \\
Linear drag $[\kdtwo]$                   & 0.1 \\
Accel.\ smooth factor $[\saccel]$        & 12.0 \\
Accel.\ smooth window $[\waccel]$ (s)    & 1.0 \\
Accel.\ ctrl delay $[\daccel]$ (s)       & 0.02 \\
Yaw smooth factor $[\syaw]$              & 12.0 \\
Yaw smooth window $[\wyaw]$ (s)          & 1.0 \\
Yaw ctrl delay $[\dyaw]$ (s)             & 0.02 \\
Vel.\ smooth factor $[\svel]$            & 0.5 \\
Vel.\ smooth window $[\wvel]$ (s)        & 2.0 \\
\midrule
\multicolumn{2}{l}{\textit{Domain randomization}} \\
Accel.\ smooth factor $[\saccel]$        & $\mathcal{U}(5,\,12)$ \\
Accel.\ ctrl delay $[\daccel]$ (s)       & $\mathcal{U}(0.01,\,0.04)$ \\
Yaw smooth factor $[\syaw]$              & $\mathcal{U}(5,\,12)$ \\
Yaw ctrl delay $[\dyaw]$ (s)             & $\mathcal{U}(0.01,\,0.05)$ \\
\bottomrule
\end{tabular}
\end{table}

\begin{table}
\centering
\caption{Nonlinear quadrotor dynamics parameters.}
\label{tab:hp-dyn-quad}
\centering
\begin{tabular}{lr}
\toprule
\textbf{Quadrotor} & \textbf{Value} \\
\midrule
Sim.\ timestep $[\dtsim]$ (s)            & 0.0025 \\
Control timestep $[\dtctrl]$ (s)         & 0.02 \\
Mass$^{\ast}$ $[\mass]$ (kg)             & 2.65 / 1.0 \\
Gravity $[\grav]$ (m/s$^{2}$)            & 9.807 \\
Quadratic drag $[\kdone]$                & 0.1 \\
Linear drag $[\kdtwo]$                  & 0.1 \\
$\Ixx,\Iyy,\Izz$ (kg\,m$^{2}$)           & 0.0143,\ 0.0175,\ 0.021 \\
$\Ixy,\Iyz,\Ixz$ (kg\,m$^{2}$)           & $2.9139 \times 10^{-7}$,\ $-4.25\times 10^{-5}$,\ $-1.91 \times 10^{-5}$ \\
Arm length $[\armlen]$ (m)         & 0.15 \\
Moment scale $[\momentscale]$                & 0.0108 \\
Motor spread angle $[\motorangle]$ (rad)        & $0.7854$ \\
Rotor time constant $[\rotortc]$ (s)     & 0.0186 \\
RPM range $[\rpmmin,\,\rpmmax]$ (rev/s)         & [2970,\ 20965] \\
\midrule
\multicolumn{2}{l}{\textit{Low-level controller}} \\
PD pos.\ gain (x,y,z)         & 6.25,\ 6.25,\ 16.0 \\
PD vel.\ gain (x,y,z)         & 5.0,\ 5.0,\ 8.0 \\
PD rot.\ gain (x,y,z)         & 110,\ 89,\ 70 \\
PD ang. vel\ gain (x,y,z)         & 20,\ 15,\ 15 \\
\bottomrule
\end{tabular}
\par\vspace{2pt}
{\footnotesize $^{\ast}$2.65~kg for \autoref{fig:algo-predef-eval}; 1.0~kg for
\autoref{fig:dyn-predef-eval}}
\end{table}


\end{document}